\documentclass[11pt,a4paper]{article}
\usepackage{emnlp2021}
\usepackage{times}
\usepackage{booktabs}
\usepackage{latexsym}
\usepackage{supertabular}
\usepackage{makecell}
\usepackage{multirow}
\usepackage{longtable}
\usepackage{subcaption}
\usepackage{enumitem}

\usepackage{microtype}

\title{On the Diversity and Limits of Human Explanations}

\author{Chenhao Tan\\
  Department of Computer Science \& Harris School of Public Policy \\
  The University of Chicago \\
  \texttt{chenhao@uchicago.edu}}

\date{}

\begin{document}
\maketitle
\begin{abstract}
A growing effort in NLP aims to build datasets of human explanations. 
However, 
it remains unclear whether these datasets serve their intended goals.
This problem is exacerbated by the fact that the term {\em explanation} is {\em overloaded} and refers to a broad range of notions with different properties and ramifications.
Our goal is to provide an overview of the diversity of explanations, discuss human limitations in providing explanations, and ultimately provide implications for collecting and using human explanations in NLP.

Inspired by prior work in psychology and cognitive sciences,
we group existing human explanations in NLP into three categories: proximal mechanism, evidence, and procedure.
These three types differ in nature and have implications for the resultant explanations.
For instance, procedure is not considered explanation in psychology and 
connects with a rich body of work on learning from instructions.
The diversity of explanations is further evidenced by proxy questions that are needed for annotators to interpret and answer ``why is [input] assigned [label]''.
Finally, giving explanations may require different, often deeper, understandings than predictions, which casts doubt on whether humans can provide valid explanations in some tasks. 
\end{abstract}

\newcommand{\viv}[1]{\textcolor{red}{[#1 ---\textsc{viv}]}}
\newcommand{\chenhao}[1]{\textcolor{blue}{[#1 ---\textsc{CT}]}}
\newcommand{\chacha}[1]{\textcolor{magenta}{[#1 ---\textsc{chacha}]}}

\newcommand{\para}[1]{\noindent{\bf #1}}
\newcommand{\figref}[1]{Fig.~\ref{#1}}
\newcommand{\tbref}[1]{Table.~\ref{#1}}
\newcommand{\secref}[1]{\S\ref{#1}}

\section{Introduction}

With the growing interest in explainable NLP systems, the NLP community have become increasingly interested in building datasets of human explanations.
These human explanations can {\em ideally} capture human reasoning of why a (correct) label is chosen.
If this is indeed the case, they are hypothesized to aid models with additional supervision, train models that explain their own predictions, and evaluate machine-generated explanations \citep{wiegreffe2021teach}.
In fact, \citet{deyoung2019eraser} already developed a leaderboard, where
the implicit assumption is that humans can provide {\em valid} explanations and these explanations can in turn be {\em uniformly} considered as groundtruths.

However, are these assumptions satisfied and can human explanations serve these goals?
In this work, we aim to introduce prior relevant literature in psychology to the NLP community and argue against abusing the term explanations and prematurely assuming that human explanations provide valid reasoning for inferring a label.

First, we point out the rich diversity in what the NLP community refer to as explanations and how researchers collect them.
The term ``explanation'' is overloaded in the NLP and AI community: it often refers to many distinct concepts and outcomes.\footnote{\citet{broniatowski2021psychological} further argues that interpretation and explainability are distinct concepts.}
For example, 
procedural instructions are different from explanations that attempt to convey proximal causal mechanisms.
The diversity of explanations is further evidenced by the variety of proxy questions that researchers ask to collect explanations, e.g., ``highlight the important words that would tell someone to see the movie'' vs. ``highlight ALL words that reflect a sentiment''.
These proxy questions are necessary because the question of ``why is [input] assigned [label]'' is too open-ended.
It follows that these ``human explanations'' are supposed to answer different questions in the first place and may not all be used for the same goals, e.g., serving as groundtruth labels.

In addition to the diversity, we highlight two insights from psychology on whether humans can provide valid explanations:
1) prediction does not entail explanation \citep{wilson1998shadows}, i.e., although humans may be able to provide valid labels, they may not be able to provide explanations that capture the reasoning process needed or used to infer a label;
2) everyday explanations are necessarily incomplete \citep{keil2006explanation,lombrozo2006structure}, because they seldom capture the complete deductive processes from a set of axioms to a statement.

\begin{table*}[t]
  \small
  \centering
  \begin{tabular}{p{0.42\textwidth}cp{0.42\textwidth}}
    \toprule
    Instance & Label & Explanation \\
    \midrule
    \multicolumn{3}{c}{Task: whether the query is supported or refuted by the preceding texts \citep{thorne-etal-2018-fever}}\\[2pt]
    \multirowcell{2}{\parbox{7cm}{\textcolor{orange}{S1:} No Way Out is the debut studio album by American hip hop recording artist , songwriter and record producer Puff Daddy . \textcolor{orange}{S2:} It was released on July 1 , 1997 , by his Bad Boy record label . The label 's official crediting as `` The Family '' 
    ...... \textcolor{orange}{Query:} 1997 was the year No Way Out was released.}} & \multirowcell{2}{\\\\Supports} & {{\bf Proximal mechanism (E1)}: ``It'' in \textcolor{orange}{S2} refers to ``No Way Out'' and ``on July 1, 1997'' is the temporal modifier of ``release'', we can thus infer 1997 was the year that No Way Out was released. \newline} \\
    &  & {\bf Evidence (E2)}: 
    {\textcolor{orange}{S1}, \textcolor{orange}{S2} }\\
    \hline
    \multicolumn{3}{c}{Task: whether \textcolor{blue}{person 1} is married to \textcolor{orange}{person 2} \citep{hancock-etal-2018-training}} \\[2pt]
    \textcolor{blue}{Tom Brady} and his wife \textcolor{orange}{Gisele Bündchen} were
    spotted in New York City on Monday amid rumors
    of Brady’s alleged role in Deflategate & \multirowcell{1}{\\True} & 
    {\bf Procedure (E3)}:  The words ``and his wife'' are  between person 1 and person 2.\\
    \bottomrule
  \end{tabular}
  \caption{Types of human explanations and corresponding examples. ``S1:'' and ``S2:'' were added to facilitate writing the explanation, which also shows the non-triviality of writing explanations.}
  \label{tab:type}
\end{table*}

In summary, {\em not} all explanations are equal and humans may {\em not} always be able to provide valid explanations.
We encourage the NLP community to embrace the complex and intriguing phenomena  behind human explanations instead of simply viewing explanations as another set of uniform labels.
A better understanding and characterization of human explanations will inform how to collect and use human explanations in NLP.

\section{Types of Human Explanations in NLP}

To understand whether datasets of human explanations can serve their intended goals, we first connect current human explanations in NLP with existing psychology literature to examine the use of the term ``explanation'' in NLP.
We adapt the categorization in \citet{lombrozo2006structure} and
group human explanations in NLP into the following three categories based on the conveyed information:

\begin{itemize} %
  \item {\bf Proximal mechanisms}.
  This type of explanation attempts to provide the mechanism behind the predicted label, i.e., {\em how} to infer the label from the text, and 
  match {\em efficient cause} in \citet{lombrozo2006structure}. 
  We created E1 in Table~\ref{tab:type} to illustrate this type of explanation. 
  Note that E1 does not provide the complete mechanism.
  For instance, it does not define ``year'' or ``temporal modifier'', or make clear that ``1997'' is a ``year''.
  Neither does it cover 
  the axioms of logic.
  This is a common property of human explanations: they 
  are known to cover {\em partial/proximal} mechanisms rather than the complete 
  deduction from natural laws and empirical conditions \citep{hempel1948studies}.
  \item {\bf Evidence}. This type of explanation
  includes the relevant tokens in the input (e.g., E2 in Table~\ref{tab:type}) and directly maps to highlights in \citet{wiegreffe2021teach}. However, it does not map to any existing definitions of explanations in the psychology literature since the evidence does not provide any information on {\em how} evidence leads to the label. 
  In other words, evidence alone does not {\em explain}.
  
  \item {\bf Procedure}. Unlike proximal mechanisms, this type of explanation provides step-by-step rules or procedures that one can directly follow, e.g., 
  E3 in Table~\ref{tab:type}. 
  They are more explicit and unambiguous than proximal mechanisms.
  In fact, one can write a rule based on E3 to find marriage relation, but one cannot easily do that with E1.
  Furthermore, the procedures are grounded to the input, so it is related to {\em formal cause}, ``the form or properties that make something what it is'' \citep{lombrozo2006structure}, which is definitional and does not convey the underlying mechanisms.
  Procedural instructions are only possible for some tasks, while proximal mechanisms are the most common form of everyday explanations.
\end{itemize}

These three categories empirically capture all the explanations discussed in NLP literature.
\citet{lombrozo2006structure} also discuss two other categories, {\em final causes} (the goal) and {\em material causes} (the constituting substance). 
For instance, a final cause to ``why [input] is assigned [label]'' can be that ``this label is provided to train a classifier''.
These two categories have been less relevant 
for NLP.

\paragraph{Implications.}
This categorization allows us to think about what kind of explanations are desired for NLP systems and help clarify how to use them appropriately.
First, proximal mechanisms are best aligned with human intuitions of explanations, especially in terms of hinting at causal mechanisms.
However, they can be difficult to collect for NLP tasks.
For example, Table~\ref{tab:esnli} shows example explanations in E-SNLI that fail to convey any proximal mechanisms: they either repeat the hypothesis or express invalid mechanisms (``the bike competition'' does not entail ``bikes on a stone road'').
See further discussions on the challenges in collecting explanations in \secref{sec:incomplete}.
Furthermore, they may be difficult to use for supervising or evaluating a model.

Second, evidence by definition provides little information about the mechanisms behind a label, but it can be potentially useful as additional supervision or groundtruths.
We will further elaborate on the nature of evidence in different tasks in \secref{sec:proxy}.
However, it may be useful to the community to use clear terminology (e.g., evidence or rationale \citep{lei-etal-2020-likely,carton+rathore+tan:20}) to avoid lumping everything into ``explanation''.

Finally, procedures are essentially instructions,
and 
\citet{keil2006explanation} explicitly
 distinguishes explanations from simple procedural knowledge:
``Knowing how to operate an automated teller machine or make an international phone call might not entail having any understanding of how either system works''.
Another reason to clarify the procedure category is that 
it would be useful to engage with a rich body of work on learning from instructions when human explanations are procedural \citep{goldwasser2014learning,matuszek2013learning}.

We would like to emphasize that procedures or instructions are powerful and can potentially benefit many NLP problems (e.g., relation extraction).
At the same time, it is useful to point out that procedures are different from proximal mechanisms.

\begin{table}
  \small
  \centering
  \begin{tabular}{p{0.45\textwidth}}
    \toprule
    \textcolor{orange}{P:} Men in green hats appear to be attending a gay pride festival.	\textcolor{orange}{H:} Men are attending a festival. \textcolor{orange}{E:}	The men are attending the festival. \\
    \midrule
    \textcolor{orange}{P:} Several bikers on a stone road, with spectators watching. \textcolor{orange}{H:}	The bikers are on a stone road.	\textcolor{orange}{E:} That there are spectators watching the bikers on a stone road implies there is a bike competition.\\
    \bottomrule
  \end{tabular}
  \caption{Examples from E-SNLI \citep{camburu2018snli}. \textcolor{orange}{P:} premise, \textcolor{orange}{H:}: hypothesis, \textcolor{orange}{E:} explanation.}
  \label{tab:esnli}
\end{table}

\begin{table*}
  \small
  \centering
  \begin{tabular}{l@{\hspace{3pt}}p{0.15\textwidth}@{\hspace{6pt}}p{0.62\textwidth}}
    \toprule
    Reference & Task & Questions \\
    \midrule
    \multicolumn{3}{c}{Evidence: single-text classification} \\
    \citet{zaidan2007using} & Sentiment analysis & {To justify why a review is positive, highlight the most important words and phrases that would tell someone to see the movie. To justify why a review is negative, highlight words and phrases that would tell someone not to see the movie.} \\
    \citet{sen2020human} & Sentiment analysis & Label the sentiment and highlight {\bf ALL} words that reflect this sentiment. \\
    \citet{carton2018extractive} & Personal attack detection & Highlight sections of comments that they considered to constitute personal attacks. \\
    \hline
    \multicolumn{3}{c}{Evidence: document-query classification} \\
    \citet{lehman2019inferring} & Question answering & Generators were also asked to provide answers and accompanying rationales to the prompts that they provided. \\
    \citet{thorne-etal-2018-fever} & Fact verification (QA) & If I was given only the selected sentences, do I have strong reason to believe the claim is true (supported) or stronger reason to believe the claim is false (refuted).
    \\
    \citet{khashabi2018looking} & Question answering &  Ask them (participants) for a correct answer and for the sentence indices required to answer the question. \\
    \bottomrule
  \end{tabular}
  \caption{Questions that prior work uses to collect human explanations. We include the short version of the guidelines here for space reasons. Refer to the appendix for the full text of the relevant annotation guidelines.}
  \label{tab:shortened_annotation}
\end{table*}

\section{Proxy Questions Used to Collect Human Explanations}
\label{sec:proxy}

Although explanations are supposed to answer 
``why is [input] assigned [label]'' \citep{wiegreffe2021teach},
this literal form 
is too open-ended and may not induce ``useful'' human explanations.
As a result, 
proxy questions are often necessary for collecting human explanations.
These proxy questions further demonstrate the diversity of human explanations beyond the types of explanation. 
Here we discuss these proxy questions for collecting evidence.
See the appendix for discussions on proximal mechanisms and procedures.
To collect evidence (highlights),
researchers adopt diverse questions 
for relatively simple single-text classification tasks (see Table~\ref{tab:shortened_annotation}).
Consider the seemingly straightforward case of sentiment analysis, ``why is the sentiment of a review positive/negative''.
A review can present both positive and negative sentiments \citep{aithal2021positivity}, so the label often comes from one sentiment outweighing the other.
However, in practice, researchers 
often 
ask annotators to identify {\bf only} words supporting the label.
Critical wording differences remain in their questions:
\citet{zaidan2007using} ask for {\em the most important} words and phrases that would {\em tell someone to see the movie},
while \citet{sen2020human} requires {\em all} words {\em reflecting the sentiment}. 
Two key differences arise: 1) ``the most important'' vs. ``all'';
2) ``telling someone to see the movie'' vs. ``reflecting the sentiment''.

In contrast, personal attack detection poses a task where the negative class (``no personal attack'') by definition points to the lack of evidence in the text.
It follows that the questions that researchers can ask almost exclusively apply to the positive class (i.e., ``highlight sections of comments that they considered to constitute personal attacks'').

In comparison, researchers approach evidence more uniformly for document-query classification tasks.
They generally use 
similar proxy questions (e.g., \citet{thorne-etal-2018-fever} and \citet{hanselowski2019richly} ask almost the same questions) and ask people to select sentences instead of words. 
That said, intriguing differences still exist:
1) \citet{lehman2019inferring} simply ask annotators to provide accompanying rationales;
2) \citet{thorne-etal-2018-fever} aim for ``strong'' reasons, which likely induces different interpretations among annotators;
3) \citet{khashabi2018looking} collect questions, answer, and sentence indices at the same time, among which sentence indices can be used to find the corresponding sentences as evidence.
It remains unclear how these differences in annotation processes and question phrasings affect the collected human explanations.

\paragraph{Implications.} Our observation on proxy questions aligns with dataset-specific designs discussed in \citet{wiegreffe2021teach}.
We emphasize that these different forms of questions entail different properties of the collected human explanations, as evidenced by \citet{carton+rathore+tan:20}.
For example, the lack of evidence in the negative class in personal attack classification likely requires special strategies in using human explanations to train a model and evaluate machine rationales.
Sentence-level and token-level annotations also lead to substantially different outcomes, at least in the forms of explanations.
We believe that it is important for the NLP community to investigate the effect of proxy questions and use the collected explanations with care, rather than lumping all datasets under the umbrella of explanations. 

We also recommend all researchers to provide detailed annotation guidelines used to collect human explanations.
As the area of collecting human explanation is nascent, the goal is not to promote consistent and uniform annotation guidelines but to encourage the community to pay attention to the different underlying questions and characterize the resultant diverse properties of human explanations.

\section{Can Humans Provide Explanations?}
\label{sec:incomplete}
In order for human explanations to serve as additional supervision in training models and evaluate machine-generated explanations, 
human explanations need to provide valid mechanisms for a correct label.
Finally, we discuss challenges for humans to provide explanations of such qualities.

\paragraph{Conceptual framework.} We situate our discussion in the psychological framework provided by \citet{wilson1998shadows} to highlight what may be required to explain. 
\citet{wilson1998shadows} examines where explanation falls in three central notions: prediction, understanding, and theories.
They argue that these three notions ``form a progression of increasing sophistication and depth with explanations falling between understanding and theories''.
For instance, we may be able to predict that a car will start when we turn the ignition switch, but few of us are able to explain in detail why this is so.
In contrast, if a person is able to explain in detail why a car starts when you turn on the ignition switch, they can likely predict what will happen if various parts of the engine are damaged or removed.

These three central notions are also essential in machine learning.
Traditional label annotation is concerned with prediction, however, being able to predict 
does not entail being able to explain.

\paragraph{Emulation vs. discovery.} Next, we gradually unfold the practical challenges in collecting valid explanations from humans. 
The first challenge 
lies in whether humans can predict, i.e., assign the correct label.
We highlight two types of tasks for AI:
emulation vs. discovery \citep{lai+liu+tan:20}.
In {\em emulation} tasks, models are trained to emulate human intelligence and labels are often crowdsourced. 
Labels, however, can also derive from external (social/biological) processes, 
e.g., the popularity of a tweet and the effect of a medical treatment.
Models can thus discover patterns that humans may not recognize 
in these {\em discovery} tasks.
While most NLP tasks such as NLI and QA are emulation tasks, many NLP problems, especially when concerned with 
social interaction, are discovery tasks, ranging from identifying memorable movie quotes to predicting the popularity of messages \citep{danescu2012you,tan+lee+pang:14}.
Aligning with our discussion on explanation and prediction, most datasets of human explanations in NLP assume that humans are able to predict and are on emulation tasks.
However, we note exceptions such as explanations of actions in gaming \citep{ehsan2019automated}, where humans may often choose sub-optimal actions (labels).

\paragraph{Cognitive challenges in providing valid explanations.} Even conditioned on that humans can predict the label, 
humans may not be able to provide valid explanations for at least two reasons. 
First, as \citet{wilson1998shadows} suggests, explanation requires more depth than prediction.
For instance, we may possess some notions of common sense (e.g., one should not slap a stranger), but it is unclear whether we can explain common sense in detail (e.g., why one should not slap a stranger through theory of morality), similar to the car ignition example.
One may argue that theory of morality may not be what NLP researchers seek, but it is critical to consider the desiderata of human explanations, with the limits in mind.

Second, explanation often requires people to report their {\em subjective} mental processes, i.e., how our minds arrive at a particular judgement, rather than following {\em objective} consensual guidelines such as annotating logical entailment.
However, classic work by \citet{nisbett1977telling} suggests that our verbal reports on our mental processes can be highly inaccurate.
For instance, in admission decisions, legitimate information can be used to justify preferences based on illegitimate factors such as race \citep{norton2006mixed}. 
Many studies on implicit bias also reinforces that we are not aware of our biases and thus cannot include them (i.e., the actual reasoning in our mind) in our explanations \citep{greenwald1998measuring}.

\paragraph{Explanations are necessarily incomplete.}
Finally, there are indeed cases where we believe that humans can provide valid mechanisms.
For instance, some question answering tasks boil down to logical inference from evidence to query.
In these cases, NLP researchers need to recognize that human explanations are necessarily incomplete: people do not start from a set of axioms and present all the deductive steps \citep{keil2006explanation,lombrozo2006structure}.
Therefore, even for simple tasks such as natural language inference, we may simply give explanations such as repeating the hypothesis without presenting %
any axiom or deduction required to infer the label.

\paragraph{Implications.} 
We cannot assume that humans are capable of providing explanations that contain valuable proximal mechanisms.
The very fact that humans can still provide explanations for incorrect labels and tasks where they do not perform well suggests that one should be skeptical about whether human explanations can be used to train models as additional supervision or evaluate machine-generated explanations as groundtruths.

Note that incomplete explanations can still be very useful for NLP. 
We believe that recognizing and characterizing this incompleteness (e.g., which proximal mechanism is more salient to humans) is critical for understanding and leveraging human explanations for the intended goals in NLP.
To summarize, we argue that human explanations are necessarily incomplete and it is important to understand and characterize this incompleteness, which can inform how we can leverage it for the intended goals in NLP.

\section{Conclusion}

Explanations represent a fascinating phenomenon and are actively studied in psychology, cognitive science, and other social sciences.
While the growing interest in explanations from the NLP community is exciting, we encourage the community to view this as an opportunity to understand how humans approach explanations and contribute to understanding and exploring the explanation processes.
This will in turn inform how to collect and use human explanations in NLP.
A modest proposal is that it is useful to examine and characterize human explanations before  assuming that all explanations are equal and chasing a leaderboard.

\section*{Acknowledgments}

We thank anonymous reviewers for their feedback, and members of the Chicago Human+AI Lab for their insightful suggestions. This work is supported in part by research awards from Amazon, IBM, Salesforce, and NSF IIS-2040989, 2125116, 2126602.

\bibliographystyle{acl_natbib}
\bibliography{refs}

\appendix

\section{Proxy Questions for Proximal Mechanisms and Procedure}

\para{Proximal mechanisms.}
In collecting proximal mechanisms, studies are more likely to ask explicitly
versions of ``why is [input] assigned [label]'', compared to the case of evidence.
However, they often need to provide structured guidelines.
For example, \citet{camburu2018snli} and \citet{rajani2019explain} discussed the need to enforce word overlap as a way to improve the quality of human rationales.
The specific requirements are quite different (see Table~\ref{tab:full_annotation_4} and Table~\ref{tab:full_annotation_5}).
There are also specific formulations of explanations, e.g., ``What aspect/stereotype/characteristic of this group (often un-fairly assumed) is referenced or implied by this post?'' in \citet{sap2019social}.
Finally, it is common that we cannot infer the exact questions asked (8/18 papers that collect explanations in free text).

\para{Procedures.}
We cannot identify the exact questions in three of five papers for explicitly step-by-step procedures, which reflects the importance of reporting detailed annotation guidelines.
As researchers collect step-by-step guidelines, \citet{ye-etal-2020-teaching} and \citet{geva2021did} adopt very different decomposition for their problems (see Table~\ref{tab:full_annotation_10}).

\section{Detailed Proxy Questions}

Table~\ref{tab:full_annotation_1}-\ref{tab:full_annotation_10} show the instructions we find in prior work that detail the proxy questions.
\citet{camburu2018snli} and \citet{rajani2019explain} collect both evidence and proximal mechanism.
We include them in the tables for proximal mechanisms.
Also, for question answering tasks, the difference between procedure and proximal mechanism can be subtle.
We consider the collected explanations {\em procedure} if they aim to explicitly provide step-by-step guides directly grounded in the input.

\clearpage
\begin{table*}
  \centering
  \begin{tabular}{lp{0.15\textwidth}p{0.6\textwidth}}
    \toprule
    Reference & Task & Questions and guidelines \\
    \midrule
    \citet{zaidan2007using} & sentiment analysis & {Each review was intended to give either a positive or a negative overall recommendation. You will be asked to justify why a review is positive or negative. {\bf To justify why a review is positive, highlight the most important words and phrases that would tell someone to see the movie. To justify why a review is negative, highlight words and phrases that would tell someone not to see the movie}. These words and phrases are called rationales. \newline
    You can highlight the rationales as you notice them, which should result in several rationales per review. Do your best to mark enough rationales to provide convincing support for the class of interest.\newline
    You do not need to go out of your way to mark everything. You are probably doing too much work if you find yourself go- ing back to a paragraph to look for even more rationales in it. Furthermore, it is perfectly acceptable to skim through sections that you feel would not contain many rationales, such as a re- viewer’s plot summary, even if that might cause you to miss a rationale here and there.} \\
    \citet{sen2020human} & sentiment analysis & {1. Read the review and decide the sentiment of this review (positive or negative). Mark your selection. \newline
    2. Highlight {\bf ALL} words that reflect this sentiment. Click on a word to highlight it. Click again to undo. \newline
    3. If multiple words refect this sentiment, please highlight them all.
    } \\
    \citet{carton2018extractive} & Personal attack detection & 40 undergraduate students used Brat (Stenetorp et al., 2012) to highlight sections of comments that they considered to constitute personal attacks. \\
    \citet{lehman2019inferring} & Question answering & {Prompt Generation: Question answering \& Prompt creators were instructed to identify a snippet, in a given full-text article, that reports a relationship between an intervention, comparator, and outcome. Generators were also asked to provide answers and accompanying rationales to the prompts that they provided; such supporting evidence is important for this task and domain. \newline The annotator was also asked to mark a snippet of text supporting their response. Annotators also had the option to mark prompts as invalid, e.g., if the prompt did not seem answerable on the basis of the article.} \\
    \citet{thorne-etal-2018-fever} & fact verification (QA) & {If I was given only the selected sentences, do I have strong reason to believe the claim is true (supported) or stronger reason to believe the claim is false (refuted). If I’m not certain, what additional information (dictionary) do I have to add to reach this conclusion. \newline
    In the annotation interface, all sentences from the introductory section of the page for the main entity of the claim and of every linked entity in those sentences were provided as a default source of evidence (left-hand side in Fig. 2). \newline
    We did not set a hard time limit for the task, but the annotators were advised not to spend more than 2-3 minutes per claim.}
    \\
    \bottomrule
  \end{tabular}
  \caption{Questions that prior work uses to solicit human explanations (\textcolor{red}{evidence}).}
  \label{tab:full_annotation_1}
\end{table*}

\begin{table*}
  \centering
  \begin{tabular}{lp{0.15\textwidth}p{0.6\textwidth}}
    \toprule
    Reference & Task & Questions and guidelines \\
    \midrule
    \citet{khashabi2018looking} & Question answering & We show each paragraph to 5 turkers and ask them to write 3-5 questions such that: (1) the question is answerable from the pas- sage, and (2) only those questions are allowed whose answer cannot be determined from a single sentence. We clarify this point by providing example paragraphs and questions. In order to encourage turkers to write meaningful questions that fit our criteria, we additionally ask them for a correct answer and for the sentence indices required to answer the question. \\
    \citet{yang2018hotpotqa} & Question answering & Workers provide the supporting facts (\underline{cannot infer the exact question})\\
    \citet{hanselowski2019richly} & Fact verification & {Stance annotation. We asked crowd workers on Amazon Mechanical Turk to annotate whether an ETS (evidence text snippets) agrees with the claim, refutes it, or has no stance towards the claim. An ETS was only con- sidered to express a stance if it explicitly referred to the claim and either expressed support for it or refuted it. In all other cases, the ETS was consid- ered as having no stance. \newline FGE annotation. We filtered out ETSs with no stance, as they do not contain supporting or refut- ing FGE. If an ETS was annotated as supporting the claim, the crowd workers selected only sup- porting sentences; if the ETS was annotated as refuting the claim, only refuting sentences were selected.}\\
    \citet{kwiatkowski2019natural} & Question answering & Long Answer Identification: For good ques- tions only, annotators select the earliest HTML bounding box containing enough information for a reader to completely infer the answer to the ques- tion. Bounding boxes can be paragraphs, tables, list items, or whole lists. Alternatively, annotators mark ‘‘no answer’’ if the page does not answer the question, or if the information is present but not contained in a single one of the allowed elements. \\
    \citet{wadden2020fact} & Fact verification & An evidence set is a collection of sentences from the abstract that provide support or contradiction for
    the given claim. To decide whether a collection of sentences is an evidence set, ask yourself, “If I were
    shown only these sentences, could I reasonably conclude that the claim is true (or false)”?
    1) Evidence sets should be minimal. If you can remove a sentence from the evidence set and the
    remaining sentences are sufficient for support / contradiction, you should remove it.
    2) There may be multiple evidence sets in a given abstract. See more at \url{https://scifact.s3-us-west-2.amazonaws.com/doc/evidence-annotation-instructions.pdf} \\
    \citet{kutlu2020annotator} & relevance assessment & Please copy and paste text 2-3 sentences from the webpage which you believe support your decision. For instance, if you selected Highly Relevant, paste some text that you feel clearly satisfies the given query. If you selected Definitely not relevant, copy and paste some text that shows that the page has nothing to do with the query. If there is no text on the page or images led you to your decision, please type ``The text did not help me with my decision''. \\
    \bottomrule
  \end{tabular}
  \caption{Questions that prior work uses to solicit human explanations (\textcolor{red}{\bf evidence}).}
  \label{tab:full_annotation_2}
\end{table*}

\begin{table*}
  \centering
  \begin{tabular}{lp{0.15\textwidth}p{0.6\textwidth}}
    \toprule
    Reference & Task & Questions and guidelines \\
    \midrule
    \citet{jansen2016s} & science QA & For each question, we create gold explanations that describe the inference needed to arrive at the correct answer. Our goal is to derive an explanation corpus that is grounded in grade-appropriate resources. Accordingly, we use two elementary study guides, a science dictionary for elementary students, and the Simple English Wiktionary as relevant corpora. For each question, we retrieve relevant sentences from these corpora and use them directly, or use small variations when necessary. If relevant sentences were not located, then these were constructed using simple, straightforward, and grade-level appropriate language. Approximately 18\% of questions required specialized domain knowledge (e.g. spatial, mathematical, or other abstract forms) that did not easily lend itself to simple verbal description, which we removed from consideration. This resulted in a total of 363 gold explanations. \\
    \citet{rajani2019explain} & Question answering & {Turkers are prompted with the following question: “Why is the predicted output the most appropriate answer?” Annotators were in- structed to highlight relevant words in the question that justifies the ground-truth answer choice and to provide a brief open-ended explanation based on the highlighted justification could serve as the commonsense reasoning behind the question. \newline Annotators cannot move forward if they do not highlight any relevant words in the question or if the length of explanations is less than 4 words. We also check that the explanation is not a sub- string of the question or the answer choices with- out any other extra words. We collect these ex- planations from only one annotator per example, so we also perform some post-collection checks to catch examples that are not caught by our previ- ous filters. We filter out explanations that could be classified as a template. For example, explanations of the form “$<$answer$>$ is the only option that is [correct|obvious]” are deleted and then reannotated.}\\
    \citet{sap2019social} & social bias & What aspect/stereotype/characteristic of this group (often unfairly assumed) is referenced or implied by this post? — Use simple phrases and do not copy paste from the post. \\
    \bottomrule
  \end{tabular}
  \caption{Questions that prior work uses to solicit human explanations for \textcolor{red}{\bf proximal mechanisms (in free text)}.}
  \label{tab:full_annotation_4}
\end{table*}

\begin{table*}
  \centering
  \begin{tabular}{lp{0.15\textwidth}p{0.6\textwidth}}
  \toprule
  Reference & Task & Questions and guidelines \\
  \midrule
  \citet{camburu2018snli} & Natual language inference & {We encouraged the annotators to focus on the non-obvious elements that induce the given relation, and not on the parts of the premise that are repeated identically in the hypothesis. For entailment, we required justifications of all the parts of the hypothesis that do not appear in the premise. For neutral and contradictory pairs, while we encouraged stating all the elements that contribute to the relation, we consider an explanation correct, if at least one element is stated. Finally, we asked the annotators to provide self-contained explanations, as opposed to sentences that would make sense only after reading the premise and hypothesis. \newline
  We did in-browser checks to ensure that each explanation contained at least three tokens and that it was not a copy of the premise or hypothesis. We further guided the annotators to provide adequate answers by asking them to proceed in two steps. First, we require them to highlight words from the premise and/or hypothesis that they consider essential for the given relation. Secondly, annotators had to formulate the explanation using the words that they highlighted. However, using exact spelling might push annotators to formulate grammatically incorrect sentences, therefore we only required half of the highlighted words to be used with the same spelling. For entailment pairs, we required at least one word in the premise to be highlighted. For contradiction pairs, we required highlighting at least one word in both the premise and the hypothesis. For neutral pairs, we only allowed highlighting words in the hypothesis, in order to strongly emphasize the asymmetry in this relation and to prevent workers from confusing the premise with the hypothesis. We believe these label-specific constraints helped in putting the annotator into the correct mindset, and additionally gave us a means to filter incorrect explanations. Finally, we also checked that the annotators used other words that were not highlighted, as we believe a correct explanation would need to articulate a link between the keywords.}
  \\
  \citet{do2020snli} & visual NLI & similar to \citet{camburu2018snli} \\
  \citet{kim2018textual} & self-driving cars & We provide a driving video and ask a human annotator in Amazon Mechanical Turk to imagine herself being a driving instructor. Note that we specifically select human annotators who are familiar with US driving rules. The annotator has to describe what the driver is doing (especially when the behavior changes) and why, from a point of view of a driving instructor. Each described action has to be accompanied with a start and end time-stamp. The annotator may stop the video, forward and backward through it while searching for the activities that are interesting and justifiable. \\
  \bottomrule
  \end{tabular}
  \caption{Questions that prior work uses to solicit human explanations for \textcolor{red}{\bf proximal mechanisms (in free text)}.}
  \label{tab:full_annotation_5}
\end{table*}

\begin{table*}
  \centering
  \begin{tabular}{lp{0.15\textwidth}p{0.6\textwidth}}
  \toprule
  Reference & Task & Questions and guidelines \\
  \midrule
  \citet{zhang2020winowhy} & coreference resolution & Given a context and a pronoun reference relationship, write how you would decide the selected candidate is more likely to be referred than the other candidate using natural language. Don't try to be overly formal, simply write what you think. In the first phase, we ask annotators to provide reasons for all WSC questions. Detailed instructions are provided such that annotators can fully understand the task1. As each question may have multiple plausible reasons, for each question, we invite five annotators to provide reasons based on their own judgments. A screenshot of the survey is shown in Figure 3. As a result, we collect 1,365 reasons. As the quality of some given reasons might not be satisfying, we introduce the second round annotation to evaluate the quality of collected reasons. In the second phase, for each reason, we invite five annotators to verify whether they think the reason is reasonable or not2. If at least four annotators think the reason is plausible, we will accept that reason. As a result, we identify 992 valid reasons. \\
  \citet{lei-etal-2020-likely} & future event prediction & we also require them to provide a rationale as to why it is more or less likely \\
  \citet{da2020edited} & harm of manipulated images & For each question we require annotators to provide both an answer to the question and a rationale (e.g. the physical change in the image edit that alludes to their answer). This is critical, as the rationales prevent models from guessing a response such as ``would be harmful'' without providing the proper reasoning for their response. We ask annotators to explicitly separate the rationale from the response by using the word ``because'' or ``since'' (however, we find that the vast majority of annotators naturally do this, without being explicitly prompted). \\
  \citet{ehsan2019automated} & gaming & ``Please explain your action''. During this time, the player’s microphone automatically turns on and the player is asked to explain their most recent action while a speech-to-text library automatically transcribes the explanation real-time. \\
  \bottomrule
  \end{tabular}
  \caption{Questions that prior work uses to solicit human explanations for \textcolor{red}{\bf proximal mechanisms (in free text)}.}
  \label{tab:full_annotation_6}
\end{table*}

\begin{table*}
  \centering
  \begin{tabular}{lp{0.15\textwidth}p{0.6\textwidth}}
  \toprule
  Reference & Task & Questions and guidelines \\
  \midrule
  \citet{ling2017program} & algebraic problems & \underline{cannot infer the exact question} \\
  \citet{alhindi-etal-2018-evidence} & fact verification & \underline{we cannot infer the exact question}  automatically extracting for each claim the justification that humans have provided in the fact-checking article associated with the claim. Most of the articles end with a summary that has a headline ``our ruling'' or ``summing up''\\
  \citet{kotonya2020explainable} & fact verification & automatically scraped from the website, \underline{we cannot infer the exact question}\\
  \citet{wang2019learning} & sentiment analysis \& relation extraction & Turkers are prompted with a list of selected predicates (see Appendix) and several examples of NL explanations. \underline{We cannot infer the exact question}\\
  \citet{brahman2020learning} & natural language inference & automatically generated. \underline{We cannot infer the exact question} \\
  \citet{li2018vqa} & visual QA & automatically generated, \underline{We cannot infer the exact question} \\
  \citet{park2018multimodal} & visual QA & During data annotation, we ask the annotators to complete the sentence “I can tell the person is doing (action) because..” where the action is the ground truth activity label. However, \underline{We cannot infer the exact question} in VQA-X.\\
\citet{rajani-etal-2020-esprit} & physics reasoning &  \underline{We cannot infer the exact question}\\
  \bottomrule
  \end{tabular}
  \caption{Questions that prior work uses to solicit human explanations for \textcolor{red}{\bf proximal mechanisms (in free text)}.}
  \label{tab:full_annotation_7}
\end{table*}

\begin{table*}
  \centering
  \begin{tabular}{lp{0.15\textwidth}p{0.6\textwidth}}
  \toprule
  Reference & Task & Questions and guidelines \\
  \midrule
  \citet{jansen-etal-2018-worldtree} & science QA & Specific interfaces were designed. For a given question, annotators identified the central concept the question was testing, as well as the inference required to correctly answer the question, then began progressively constructing the explanation graph. Sentences in the graph were added by querying the tablestore based on key- words, which retrieved both single sentences/table rows, as well as entire explanations that had been previously annotated. If any knowledge required to build an explanation did not exist in the table store, this was added to an appropriate table, then added to the explanation. \\
  \citet{xie-etal-2020-worldtree} & science QA & similar to \citet{jansen-etal-2018-worldtree}\\
  \citet{khot2020qasc} & question answering & The HIT here is to write a test question that requires CHAINING two facts (a science fact and some other fact) to be combined. \\
  \citet{jhamtani-clark-2020-learning} & question answering & We then use (Amazon Turk) crowdworkers to annotate each chain. Workers were shown the question, correct answer, and reasoning chain. They were then asked if fact 1 and fact 2 together were a reasonable chain of reasoning for the answer, and to promote thought were offered several categories of ``no'' answer: fact 1 alone, or fact 2 alone, or either alone, justified the answer; or the answer was not justified; or the question/answer did not make sense. (Detailed instructions in the appendix)\\
  \citet{inoue-etal-2020-r4c} & question answering & 1. Read a given question and related articles. 2. Answer to the question solely based on the information from each article. 3. Describe your reasoning on how to reach the answer. Each reasoning step needs to be in a simple subject-verb-object form (see example below). Your reasoning must include sentences containing your answer.\\
  \bottomrule
  \end{tabular}
  \caption{Questions that prior work uses to solicit human explanations in \textcolor{red}{\bf proximal mechanisms (in structured explanations)}.}
  \label{tab:full_annotation_8}
\end{table*}

\begin{table*}
  \centering
  \begin{tabular}{lp{0.15\textwidth}p{0.6\textwidth}}
  \toprule
  Reference & Task & Questions and guidelines \\
  \midrule
  \citet{srivastava-etal-2017-joint} & concept learning & The screenshot includes both ``explanations'' and ``instructions'', however, \underline{we cannot infer the exact question} \\
  \citet{hancock-etal-2018-training} & relation extraction &  \underline{we cannot infer the exact question}\\
  \bottomrule
  \end{tabular}
  \caption{Questions that prior work uses to solicit human explanations \textcolor{red}{\bf for procedure (in free text)}.}
  \label{tab:full_annotation_9}
\end{table*}

\begin{table*}
  \centering
  \begin{tabular}{lp{0.15\textwidth}p{0.6\textwidth}}
  \toprule
  Reference & Task & Questions and guidelines \\
  \midrule
  \citet{lamm2020qed} & question answering & referential equality, \underline{we cannot infer the exact question} \\
  \citet{ye-etal-2020-teaching} & question answering & Please read carefully to get accepted!
  (1) You're not required to answer the question. The answer is already provided and marked in red. Read examples below carefully to learn about what we want!
  (2) Identify important short phrases that appear both in the question and in the context.
  Important: The two appearances of the phrase should be exactly the same (trivial differences like plural form or past tense are still acceptable).
  Important: Write sentences like Y is "Switzerland". Make sure there is no typo in what you quote.
  (3) Explain how you locate the answer with the phrases you marked; Only use the suggested expressions in the table in the bottom.\\
  \citet{geva2021did} & question answering & 1)  Creative question writing: Given a term (e.g., silk), a description of the term, and an expected answer (yes or no), the task is to write a strategy question about the term with the expected answer, and the facts required to answer the question. 2) Strategy question decomposition: Given a strategy question, a yes/no answer, and a set of facts, the task is to write the steps needed to answer the question. 3) Evidence matching: Given a question and its de- composition (a list of single-step questions), the task is to find evidence paragraphs on Wikipedia for each retrieval step. Operation steps that do not require retrieval are marked as operation. \\
  \bottomrule
  \end{tabular}
  \caption{Questions that prior work uses to solicit human explanations \textcolor{red}{\bf for procedure (in structured explanations)}.}
  \label{tab:full_annotation_10}
\end{table*}

\clearpage

\end{document}